\documentclass[letterpaper]{article} 
\usepackage[]{aaai2027}  
\usepackage[hyphens]{url}  
\usepackage{graphicx} 
\usepackage{natbib}  
\usepackage{caption} 
\usepackage{algorithm}
\usepackage{algorithmic}

\usepackage{multirow} 
\usepackage{newfloat}
\usepackage{listings}
\DeclareCaptionStyle{ruled}{labelfont=normalfont,labelsep=colon,strut=off} 
\floatstyle{ruled}
\newfloat{listing}{tb}{lst}{}
\floatname{listing}{Listing}

\usepackage{booktabs}
\usepackage{amsfonts}
\usepackage{amssymb}
\usepackage{amsmath}
\title{Measure, Don't Optimize: Forecasting Recovery in LLM Unlearning}
\author{
    Zirui Song\textsuperscript{\rm 1,\rm 2}\thanks{Work done during internship at AMAP, Alibaba Group.},
    Huaxing Liu\textsuperscript{\rm 2},
    Xiang Wang\textsuperscript{\rm 2},
    Shuai Li\textsuperscript{\rm 2},
    Xinye Li\textsuperscript{\rm 1},
    Lang Gao\textsuperscript{\rm 1},\\
    Jinghui Zhang\textsuperscript{\rm 1},
    Zheng Lu\textsuperscript{\rm 3},
    Fengxian Ji\textsuperscript{\rm 1},
    Xiaojun Chang\textsuperscript{\rm 1},
    Xiuying Chen\textsuperscript{\rm 1}\thanks{Corresponding author.}
}
\affiliations{
    \textsuperscript{\rm 1}Mohamed bin Zayed University of Artificial Intelligence (MBZUAI)\\
    \textsuperscript{\rm 2}AMAP, Alibaba Group, \\
    \textsuperscript{\rm 3}Peking University, \\
}

\newcommand{\jocc}{\textsc{J-Access}}
\newcommand{\joccs}{\mathrm{JOcc}}

\begin{document}

\maketitle

\begin{abstract}
Prior white-box studies show that large language models can retain latent traces of target knowledge after unlearning, even when the knowledge is no longer expressed in their outputs.
However, existing audits remain limited to one-off diagnostics: it is unclear whether these residual signals can predict future recovery under continued training or serve as reliable optimization targets. Resolving this gap is essential to determine whether internal auditing can move beyond post-hoc evaluation toward proactive risk monitoring and safer unlearning.
We propose \jocc{}, an inference-time audit that uses the Jacobian lens to map intermediate representations into vocabulary space and measures how often target concepts remain accessible along the model’s output pathway. 
We hypothesize that residual accessibility reflects recovery susceptibility: knowledge that remains closer to the output pathway requires less fine-tuning to restore, leading to faster recovery.
We audit 398 public unlearned models spanning eight unlearning methods. We find that: (1) most unlearned models retain access above the retain-only gold level. (2) pre-attack accessibility predicts recovery speed and extent at the model level, but cannot identify which specific facts will be recovered; and (3) directly minimizing \jocc{} does not promote genuine deletion. Instead, the model learns to hide knowledge from the audit, producing lower audit scores but greater post-attack recovery.
These findings position \jocc{} as a model-level diagnostic for assessing residual susceptibility in unlearned models.
We argue internal audits should serve as an independent diagnostic dimension in unlearning evaluation, and should not be converted into optimization targets without validation.

\end{abstract}

\section{Introduction}
A growing body of white-box research has found that large language models (LLMs) can retain traces of "forgotten" knowledge in their hidden states and parameters after undergoing machine unlearning, even when that knowledge no longer appears in the model's output \citep{patil2024can,lynch2024eight,hong2024dissecting,hong2025intrinsic,lee2026measuring,song2026audio,wang2025word}. This finding challenges the dominant practice in unlearning evaluation: whether an unlearning method succeeds is typically certified purely at the behavioral level, by measuring answer probabilities, text overlap, truth ratios, question-answering performance, and membership-inference leakage on the forget set \citep{maini2024tofu,shi2025muse,cao2024rwku,dorna2026openunlearning,gao2025evaluate}. 
We refer to success under such output-based evaluations as behavioral forgetting~\cite{yang2026distinguishable,song-etal-2025-injecting,lee2026measuring}. These results indicate that what a model expresses does not always reflect what it retains, undermining the assumption on which behavioral forgetting is typically treated as equivalent to true deletion.

\begin{figure}[t]
\centering
\includegraphics[
trim=7 7 0 7, 
clip, 
width=\linewidth,
]{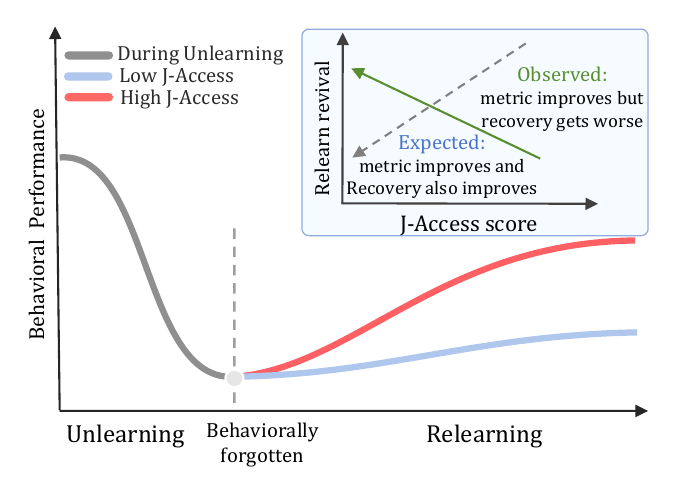}
\caption{\textbf{J-Access predicts future recovery risk but fails as a direct optimization target.}
Higher pre-attack J-Access predicts greater model-level recovery, whereas directly minimizing it can lower the audit score without deleting the underlying knowledge.
}
\label{fig:overview}
\end{figure}

However, detecting a residual trace alone does not establish its practical significance, because existing white-box audits provide only a single-point-in-time snapshot. This limitation raises two practical questions. First, can residual signals predict future knowledge recovery under continued training? Such predictive validity is necessary to distinguish temporary suppression from stable forgetting, compare the long-term risks posed by different unlearning methods, and determine whether the signal is reliable at the model or instance level. Second, do these signals remain reliable when directly optimized? This robustness is essential if internal audits are to guide training, since a model may learn to suppress the measured signal without genuinely deleting the underlying knowledge. As summarized in Figure~\ref{fig:overview}, we evaluate internal auditing in two settings: as an independent predictor of future recovery and as a direct optimization target during training.

To answer these questions, we propose \jocc{}, an inference-time auditing method that measures the internal distance between target knowledge and the model's output pathway, using the Jacobian~\cite{jacobi1841determinantibus} lens to operationalize this notion.
Specifically, for each probe query that implicates a forgotten entity without naming it, we map mid-depth residual representations from a band of mid-to-late layers into vocabulary space via the Jacobian lens, and check whether any token associated with the target concept appears among the top-ranked decoded tokens; the resulting access rate is normalized against a gold retain-only model and the original pre-unlearned model. 
This readout mechanism builds on recent interpretability work identifying such mid-depth representations as a functional "workspace" for verbalizable content, and causally validating the readout via steering and patching experiments \citep{gurnee2026verbalizable}.
This design is naturally suited to answering whether future recovery can be predicted: relearning attacks essentially reconnect, via a small amount of fine-tuning, knowledge that already exists but is temporarily disconnected from the output pathway. The closer that knowledge sits to the output pathway, the less fine-tuning is needed to reopen it, and the faster and more completely recovery occurs.

We conduct experiments on the TOFU benchmark \citep{maini2024tofu} and  OpenUnlearning \citep{dorna2026openunlearning}, spanning eight unlearning methods and 398 public unlearned models. We report three main findings. 
First, most unlearned models, including those passing standard forgetting and utility criteria, retain internal access to the target knowledge well above the retain-only gold level, showing that behavioral forgetting does not imply internal erasure.
Second, across cross-entity relearning attacks, pre-attack \jocc{} predicts model-level recovery speed, but not which specific items recover. 
Third, directly suppressing the \jocc{} score during unlearning lowers the audit score but increases post-attack recovery, indicating that the model learns to evade the audit rather than genuinely delete the target knowledge.

\section{Related Work}
\label{sec:related}

\paragraph{LLM unlearning methods.}
Machine unlearning originates in exact retraining schemes such as
SISA \citep{bourtoule2020machineunlearning} and certified removal guarantees
\citep{guo2023certi}, which do not scale to LLM training runs, so
approximate objectives dominate. Optimization-based unlearning ascends the loss on the forget set\citep{jang2023knowledge} and stabilizes it with retain-side regularization, as in GradDiff \citep{maini2024tofu}. Preference-style objectives such as NPO, SimNPO, and AltPO reframe forgetting as rejecting or replacing the target answers\citep{zhang2024negative,fan2026simplicity,mekala2025alternate}; refusal-target training maps forget queries to abstention \citep{maini2024tofu}; UNDIAL self-distills adjusted logits \citep{dong2025undial}; and RMU perturbs intermediate representations on hazardous data \citep{li2024wmdp}. \citet{eldan2023s} pioneered approximate unlearning of pretrained content, and \citet{liu2025rethinking} survey this rapidly growing space. We audit released checkpoints of eight such methods and ask what their behavioral success conceals.

\paragraph{Behavioral evaluation of unlearning.} 
TOFU scores forgetting against a gold retain-only reference on fictitious authors \citep{maini2024tofu}; MUSE evaluates verbatim memorization, knowledge, privacy leakage, and utility \citep{shi2025muse}; WMDP measures hazardous capabilities via multiple choice \citep{li2024wmdp}; RWKU extends evaluation to real-world entities and adversarial prompts \citep{cao2024rwku}. OpenUnlearning consolidates methods and metrics, including the membership-inference attacks we use as baselines, and releases the unlearned models we audit \citep{dorna2026openunlearning}. 
Recent critiques argue that such benchmarks are weak measures of progress because suppression can masquerade as forgetting\citep{thaker2025position,gao2025evaluate}.
All of these certify unlearning through model outputs; None examines whether the target knowledge remains internally accessible, or whether such residual access carries any information about future recovery.

\paragraph{White-box audits of unlearned models.}
Prior work shows that behavioral forgetting does not imply internal erasure. Residual knowledge can be detected in hidden states \citep{patil2024can,lynch2024eight}, localized through activation patching or parameter restoration \citep{hong2024dissecting}, quantified through parametric traces \citep{hong2025intrinsic}, and recovered through activation steering \citep{seyitouglu2024extracting}. Closest to our work, UDS measures internal erasure using two-stage activation patching \citep{lee2026measuring}. However, these audits are validated only against behavior at the same checkpoint. Whether their signals can predict future relearning or remain reliable under direct optimization is still unknown. This concern is supported by evidence that latent-space defenses can be evaded through activation obfuscation \citep{bailey2025obf}, consistent with the broader vulnerability of proxy objectives to over-optimization \citep{gao2022scaling}.

To address these gaps, we decode residual representations directly into vocabulary space, building on the logit lens and its refinements \citep{belrose2023eliciting,ghandeharioun2024patchscopes}. Specifically, we adopt the Jacobian lens \citep{gurnee2026verbalizable}, because unlearning may shift representations and violate the shared-basis assumption underlying direct unembedding. Related readouts have also revealed latent knowledge that models do not explicitly verbalize \citep{burns2022discovering,ryd2025towards}. Using this readout, we evaluate white-box auditing along two previously unexplored dimensions: prospective validity and robustness under direct optimization.

\paragraph{Relearning and other recovery attacks.} 
Few-epoch fine-tuning recovers ``unlearned'' knowledge \citep{hu2024jogging,deeb2024unlearning,lucki2024adversarial}; forgotten content can also be reintroduced in context \citep{shumailov2024ununlearning}, and low-bit quantization alone restores much of it by erasing the small weight updates that unlearning applies \citep{zhang2025catastrophic}. Recent methods aim to resist such attacks \citep{fan2025towards,guo2024mechanistic}. Prior work uses these attacks to demonstrate that specific methods fail; we instead use relearning, with attack rules and success criteria fixed in advance, as the validation standard against which an internal audit's predictions are scored.

\begin{figure*}[t]
\centering
\includegraphics[width=\textwidth]{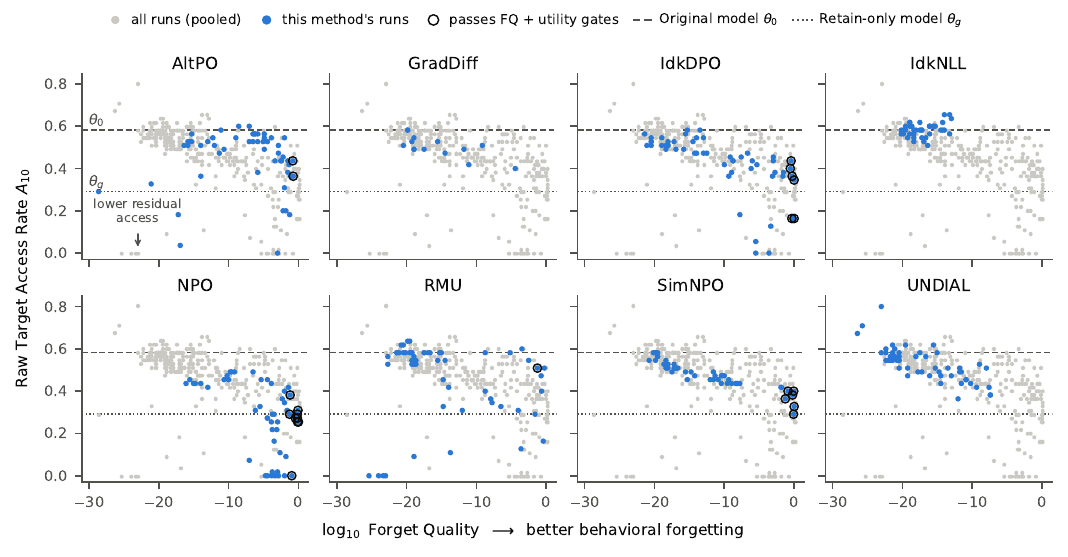}%
\hfill
\caption{\textbf{Behavioral forgetting does not imply internal erasure.}
Across 398 unlearned models, most retain higher target accessibility than the retain-only gold model, including many behaviorally successful checkpoints. J-Access also varies substantially across and within unlearning methods. Each panel highlights one method in blue; circled points indicate behavioral success, while dashed and dotted lines denote the original and gold models, respectively.}
\label{fig:audit}
\end{figure*}

\section{The J-Access Audit}
\label{sec:background}

\subsection{Notation}
\label{sec:data}

A model with parameters $\theta_0$ is trained on
$\mathcal{D}=\mathcal{D}_r\cup\mathcal{D}_f$, where $\mathcal{D}_f$ is
the forget set and $\mathcal{D}_r$ the retain set; an unlearning
algorithm maps $\theta_0\!\to\!\theta_u$ using
$\mathcal{D}_f,\mathcal{D}_r$; the gold reference $\theta_g$ is
trained on $\mathcal{D}_r$ alone. An audit probe $x_i$ is a query
that implicates a forget-set entity without containing the target
answer; its concept token set $C_i$ collects the tokens that
would express the target association. We use two granularities: the primary identity level, the full-name and name-piece tokens of the implicated entity, and a secondary knowledge level, which extends $C_i$ to answer content words under document-frequency and inferability filters, details in Appendix A.

\subsection{The Jacobian Lens}
\label{sec:lens}

Our audit asks whether a target association remains available in a
model's intermediate representations even when the model does not
express that association in its output. 
\textcolor{black}{To measure this form of accessibility, we require a readout that estimates how an intermediate representation would be transformed by the model's subsequent layers into vocabulary-level evidence. Directly applying the unembedding matrix to an intermediate state, as in the logit lens, assumes that intermediate and final-layer representations share a common basis. This assumption may be unreliable after unlearning, which can alter the geometry of the residual stream without removing the underlying association. We therefore use the Jacobian lens~\citep{gurnee2026verbalizable}, which transports intermediate representations through a linear approximation of the model's downstream computation before decoding them into vocabulary space.}

Consider a decoder-only transformer with $L$ layers, hidden dimension
$d$, vocabulary $\mathcal{V}$, and unembedding matrix
$W_U\in\mathbb{R}^{|\mathcal{V}|\times d}$. For an input $x$, let
$h_{\ell,p}(x)\in\mathbb{R}^{d}$ denote the residual-stream state at
layer $\ell$ and token position $p$.

A direct application of the model's unembedding matrix to
$h_{\ell,p}$ assumes that intermediate and final-layer representations
share the same basis. This assumption need not hold. The Jacobian lens instead
maps an intermediate state into the basis of a later target layer
through a corpus-averaged Jacobian:
\begin{equation}
J_\ell =
\mathbb{E}_{x,p,p'\ge p}
\left[
\frac{\partial h_{T,p'}(x)}
     {\partial h_{\ell,p}(x)}
\right],
\label{eq:jacobian}
\end{equation}
where $T$ is the target layer. The expectation is taken over generic
text inputs, source positions $p$, and current or future target
positions $p'\ge p$. The restriction $p'\ge p$ respects the causal
structure of decoder-only language models.

The transported state is decoded into vocabulary space using the
model's output normalization and unembedding:
\begin{equation}
z_{\ell,p}(x)
=
W_U\,
\operatorname{Norm}
\left(
J_\ell h_{\ell,p}(x)
\right)
\in\mathbb{R}^{|\mathcal{V}|}.
\label{eq:lens}
\end{equation}
Each coordinate of $z_{\ell,p}(x)$ corresponds to one vocabulary
token. We write
$\operatorname{rank}_{t}[z_{\ell,p}(x)]$ for the rank of token $t$
after sorting these coordinates in decreasing order, with rank 1
indicating the token most strongly supported by the readout.

Prior work shows that the Jacobian lens can reveal verbalizable concepts in intermediate representations before those concepts appear in the model's output, and causally validates this readout through steering, ablation, and representation-swapping experiments on general text understanding tasks \citep{gurnee2026verbalizable}. 
We repurpose this validated readout for a setting it was not designed for: auditing whether an unlearned association remains internally accessible even when the model has been explicitly trained to suppress its expression, a question the original readout was never tested against.

\subsection{The J-Access Score}
\label{sec:jocc}

With the Jacobian-lens readout in place, we now convert its token-level evidence into a normalized measure of residual accessibility. We propose J-Access, an inference-time audit that searches a predefined band of intermediate representations for target-concept evidence and aggregates the results across probes. The resulting access rate is normalized between the original model and a retain-only gold model, enabling comparisons across unlearned checkpoints.

Let $x_1,\ldots,x_N$ denote the audit probes. Each probe $x_i$ implicates one forget-set association without explicitly containing its target answer, and $C_i\subset\mathcal{V}$ denotes the corresponding set of concept tokens.

For model $\theta$, we apply the Jacobian lens to residual states within a
predefined workspace band $\mathcal{B}$ and a set of readout positions
$\mathcal{P}$. We say that the target association is accessible if at
least one target token enters the Jacobian lens top-$k$ at any audited layer
and position:
\begin{equation}
\begin{split}
a_i^{(k)}(\theta)
= \mathbb{1}\Big[
&\ \exists\, \ell\in\mathcal{B},\ p\in\mathcal{P},\ t\in C_i \\
&\ \text{s.t.}\ 
\operatorname{rank}_{t}\!\left[ z_{\ell,p}(x_i;\theta) \right]
\le k
\Big].
\end{split}
\label{eq:access-indicator}
\end{equation}
Thus, $a_i^{(k)}(\theta)=1$ indicates that some token associated with
the target concept appears among the $k$ highest-scoring vocabulary
tokens at least once within the audited workspace.

Our goal is specifically to measure access without expression.
We therefore restrict the audit to probes on which the model does not
behaviorally produce the target concept:
\begin{equation}
\mathcal{I}(\theta)
=
\left\{
i:
\text{$\theta$ does not emit any token in $C_i$}
\right\}.
\label{eq:silent-set}
\end{equation}
Expressed cases are handled by standard behavioral metrics rather than
by the internal audit.

For any model $\phi$ and probe subset $\mathcal{I}$, we define the raw
access rate as
\begin{equation}
A_k(\phi;\mathcal{I})
=
\textstyle \frac{1}{|\mathcal{I}|}
\sum_{i\in\mathcal{I}}
a_i^{(k)}(\phi).
\label{eq:raw-access}
\end{equation}
This quantity measures the fraction of behaviorally silent probes for
which the target association nevertheless appears among the Jacobian lens
top-$k$ tokens.

Because raw access rates depend on both the accessibility of the
original model and the nonzero background rate of a model that never
learned the target data, we normalize each audited checkpoint between
two empirical anchors. Let $\theta_0$ denote the original model trained
on the forget data and $\theta_g$ the gold retain-only model. Both
anchors are evaluated on the same silent-probe subset
$\mathcal{I}(\theta)$ selected for the audited checkpoint:
\begin{equation}
\joccs(\theta)
=
\frac{
A_k\!\left(\theta;\mathcal{I}(\theta)\right)
-
A_k\!\left(\theta_g;\mathcal{I}(\theta)\right)
}{
A_k\!\left(\theta_0;\mathcal{I}(\theta)\right)
-
A_k\!\left(\theta_g;\mathcal{I}(\theta)\right)
}.
\label{eq:jaccess}
\end{equation}

Under this normalization, $\joccs(\theta)=1$ indicates the same level of measured accessibility as the original model, whereas
$\joccs(\theta)=0$ indicates accessibility at the retain-only gold level. Accordingly, $1-\joccs(\theta)$ represents the fraction of the measured internal distance from the original model toward the gold model traversed by the unlearned checkpoint. Scores are not clipped, so values outside $[0,1]$ indicate that a checkpoint lies beyond one of the two empirical anchors.

In the primary analysis, we use identity-level concept tokens, $k=10$, and a mid-to-late workspace band selected on held-out calibration entities. The construction of concept sets, readout positions, and layer selection is described in Appendix~A of the supplementary material; robustness variants are reported in Appendix~B.

\begin{table}[t]
\centering
\small
\begin{tabular}{lrrr}
\toprule
Method & $>$gold & sig. & med.\ $\jocc{}$ \\
\midrule
AltPO     &  85\% &  67\% & 0.75 \\
GradDiff  &  85\% &  68\% & 0.62  \\
IdkDPO    &  87\% &  78\% & 0.59  \\
IdkNLL    & 100\% & 100\% & 1.00  \\
NPO       &  46\% &  26\% & 0.00  \\
RMU       &  83\% &  67\% & 0.88  \\
SimNPO    &  98\% &  88\% & 0.62  \\
UNDIAL    & 100\% &  96\% & 0.88  \\
\midrule
All     &  85\% &  73\% & 0.69\\
\bottomrule
\end{tabular}
\caption{Residual accessibility, grouped by unlearning method. $>$gold denotes the percentage of unlearned models whose raw target access exceeds the retain-only gold anchor; sig. denotes the percentage significantly above the gold anchor after Holm correction over the full grid; median \jocc{} denotes the median normalized audit score within each method.}
\label{tab:audit}
\end{table}

\section{Experimental Setup}
\label{sec:setup}

We evaluate \jocc{} through three progressively stronger tests.
We first ask whether it reveals residual access that is not captured by behavioral evaluation. We then ask whether this residual signal prospectively predicts recovery under relearning. Finally, we test whether the audit remains informative when it is itself placed under optimization pressure.

\subsection{Benchmark, Models and Behavioral Metric}

We use TOFU \texttt{forget10}~\citep{maini2024tofu} as our primary benchmark. TOFU provides a precisely specified forget set consisting of question--answer pairs about fictitious authors, together with a retain-only reference model $\theta_g$ trained without the forget authors.

For the full-grid audit, we evaluate all 398 unlearned models released by OpenUnlearning~\citep{dorna2026openunlearning}, spanning GradDiff~\cite{maini2024tofu}, NPO~\cite{zhang2024negative}, SimNPO~\cite{fan2026simplicity}, AltPO~\cite{mekala2025alternate}, IdkDPO~\cite{maini2024tofu}, IdkNLL~\cite{maini2024tofu}, UNDIAL~\cite{dong2025undial}, and RMU~\cite{li2024wmdp}. The subsequent relearning study uses a stratified subset of these models, while the optimization study trains dedicated audit-suppression variants.

We follow the official TOFU and OpenUnlearning behavioral evaluation, including Forget Quality (FQ), Model Utility (MU), answer probability, ROUGE overlap, and truth ratio \citep{maini2024tofu,dorna2026openunlearning}. 

\subsection{Test 1: Residual Access}

We first ask whether \jocc{} captures variation that is invisible to behavioral evaluation. Rather than selecting a small set of representative models, we audit the complete OpenUnlearning of 398 unlearned models. 
We test whether each unlearned models exhibits greater raw access than the gold anchor, correcting across the full grid using Holm's procedure.

We next test whether it behaves as a meaningful measure of residual access rather than an artifact of the readout. We refit the lens after unlearning to account for possible representation shift and compare \jocc{} with the Unlearning DepthScore (UDS)~\citep{lee2026measuring}.
Because UDS estimates deletion through activation patching rather than vocabulary-space decoding, this comparison tests convergent validity across two mechanistically distinct measurements.

\subsection{Test 2: Recovery Prediction}
We evaluate whether pre-attack \jocc{} predicts subsequent recovery stratified by unlearning method and forgetting quality. Following prior work \citep{hu2024jogging,deeb2024unlearning,fan2025towards}, each unlearned model is fine-tuned on question--answer pairs from a subset of forgotten entities and evaluated on disjoint held-out entities. The gold model $\theta_g$ undergoes the identical attack, providing a reference for gains attributable to fresh learning rather than revival under the same fine-tuning procedure.

We measure recovery with two outcomes. An item counts as revived if its post-attack ROUGE recall reaches half of the original model's score, having been below this threshold before the attack; excess revival is a checkpoint's revival rate minus that of the identically attacked gold model. Steps-to-recover is the first attack step at which the mean answer probability on held-out items reaches half of the original model's level, censored at the attack horizon when never reached within the predefined attack budget.

At the model level, we report Spearman correlations between pre-attack \jocc{} and these outcomes over the full pool, as a partial rank correlation controlling for Forget Quality and Model Utility, and within method families, where the eight within-method coefficients are pooled by inverse-variance weighting after Fisher transformation. The knowledge-level variant repeats the pooled analysis with \jocc{} computed on the extended concept-token set, keeping excess revival as the outcome.

At the item level, we test whether pre-attack signals distinguish items that revive from those that remain suppressed, excluding items recovered before the attack. Each item is scored by the highest rank attained by its concept tokens within the workspace band, and performance is measured by the median per-checkpoint AUROC. We compare \jocc{} with TOFU answer probability, six membership-inference attacks from OpenUnlearning, and a logit-lens baseline that decodes the same representations without Jacobian transport. We further evaluate the incremental value of \jocc{} by adding it to a logistic regression over the baseline predictors and measuring the change in held-out AUROC under checkpoint-grouped cross-validation. Full predictor definitions and attack hyperparameters are provided in Appendix~C.

\subsection{Test 3: J-Access as an Unlearning Objective}

We directly optimize the audit using WD-Train, which augments the unlearning objective with a penalty on Jacobian-lens accessibility. We evaluate suppression weights $\lambda\in\{0,5,10\}$ using three random seeds per setting. 
The differentiable suppression objective, training hyperparameters, and seed-level results are provided in Appendix~D.

For each setting, we measure \jocc{}, causal deletion depth using UDS, Model Utility, and post-attack revival. We further compare WD-Train with GradDiff and RMU configurations exhibiting shallow or deep deletion under UDS. These controls distinguish genuine deletion, which should reduce both accessibility and recovery, from suppression that only lowers the audit score.

\begin{figure}[t]
\centering
\includegraphics[width=\linewidth]{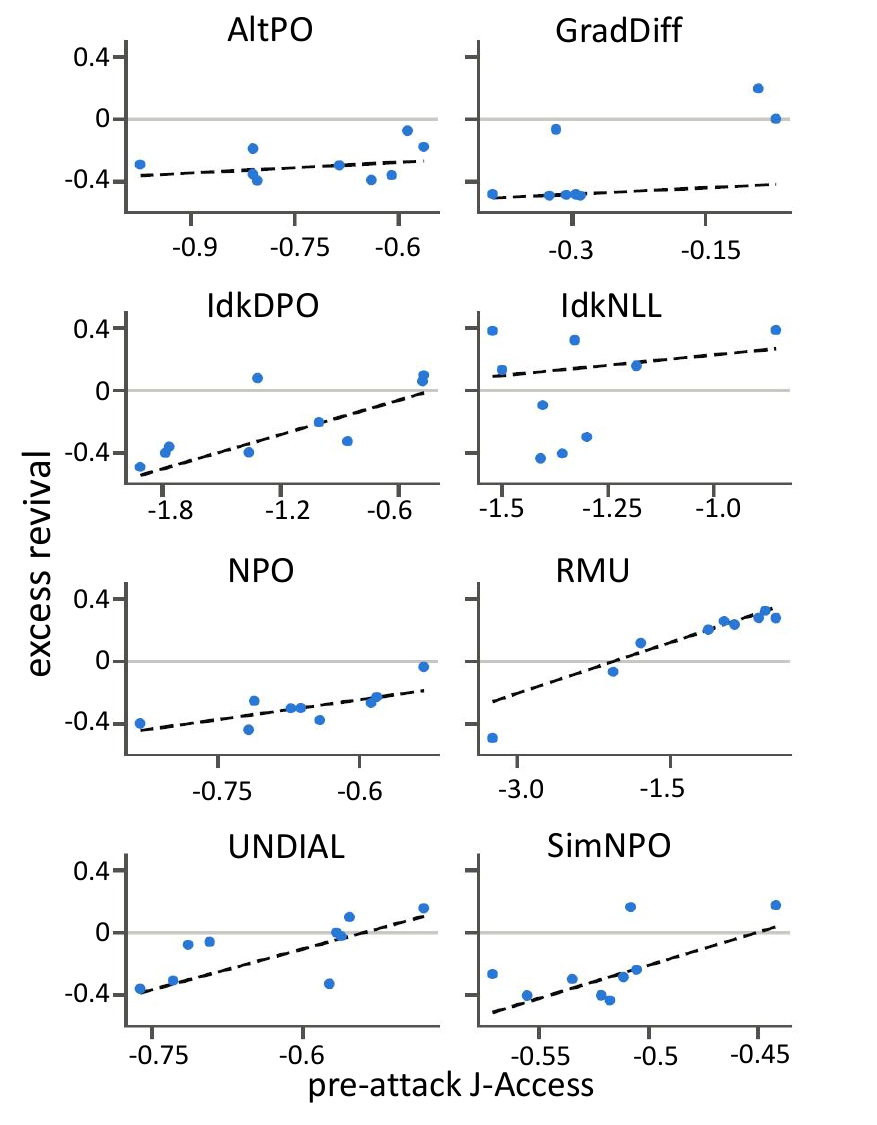}
\caption{\textbf{Higher pre-attack \jocc{} predicts greater recovery within each unlearning method.}
Each point represents an unlearned model, with excess revival measured on held-out entities. Dashed lines show Theil--Sen fits, indicating that the association is not driven solely by differences between methods.}
\label{fig:predict}
\end{figure}

\begin{table}[t]
\centering
\small
\begin{tabular}{lr}
\toprule
Model-level analysis & $\rho$  \\
\midrule
Full pool, excess revival        & $+0.35$ \\
\quad partial (FQ, MU controlled)& $+0.33$ \\
\quad within-method pooled       & $+0.45$ \\
\quad knowledge-level variant  & $+0.71$ \\
Steps-to-recover                 & $-0.70$  \\
Behaviorally successful subset  & $+0.18$  \\
\bottomrule
\end{tabular}
\caption{\textbf{Pre-attack \jocc{} predicts model-level recovery under relearning.}
Entries report Spearman correlations with excess revival or steps to recovery; bold values are statistically significant ($p<0.05$). The partial estimate controls for Forget Quality (FQ) and Model Utility (MU), while the within-method estimate pools correlations across unlearning methods. The knowledge-level row uses the extended concept-token set, and the behaviorally successful subset includes checkpoints with Forget Quality $>0.01$.}
\label{tab:predict}
\end{table}

\section{Results}

\subsection{Test 1: Does Behavioral Forgetting Ensure Erasure?}
\label{sec:audit}
 
Unlearned models retain widespread internal access to the forgotten knowledge. Across all 398 unlearned models, 85\% exhibit higher target access than the retain-only gold model, and the median normalized accessibility is $0.69$ (Table~\ref{tab:audit}). Thus, the typical unlearned model closes less than one third of the measured internal gap between the original and gold models. 
As shown in Fig.~\ref{fig:audit}, unlearned models with similar behavioral scores also span a wide range of accessibility values, both across and  within method families. Behavioral forgetting therefore constrains, but does not determine, residual internal accessibility.

This variation has two implications. First, behavioral metrics cannot reliably rank checkpoints by internal erasure, because models with comparable Forget Quality may occupy substantially different positions between the original and gold anchors. Second, the substantial within-method variation shows that residual access depends not only on the unlearning algorithm but also on its training configuration and checkpoint selection. Evaluating a single representative checkpoint may therefore obscure meaningful differences in deletion quality. \jocc{} complements behavioral metrics by revealing this hidden variation across otherwise behaviorally similar models.

The measured signal is absent for unseen twin entities, exceeds the matched-decoy baseline, and remains stable when the lens is refitted after unlearning. Moreover, \jocc{} agrees with the independent activation-patching criterion in the expected direction: unlearned models with greater residual access exhibit shallower causal deletion the Spearman $\rho=-0.72$. Full statistics for the individual controls and robustness variants are reported in Appendix~B. Together, these results indicate that \jocc{} captures target-specific residual access rather than a generic readout artifact.


\begin{table}[t]
\centering
\small
\begin{tabular}{lr}
\toprule
Item-level predictor &  AUROC \\
\midrule
Pre-attack probability      & 0.582 \\
MIA (best of six)           & 0.579 \\
\jocc{} (knowledge level)   & 0.555 \\
Logit-lens accessibility        & 0.548 \\
\jocc{} (preregistered) & 0.504 \\
\midrule
Stacked $\Delta$AUC of \jocc{} over behavior+MIA & $+0.0002$ \\
\bottomrule
\end{tabular}
\caption{\textbf{\jocc{} does not reliably predict which individual facts will recover.}
Item-level AUROC remains near chance ($0.5$), and adding \jocc{} to behavioral and membership-inference predictors yields negligible improvement. The preregistered row uses the knowledge-level score with its orientation fixed in advance.}
\label{tab:item}
\end{table}

\subsection{Test 2: Does J-Access Predict Future Recovery?}
\label{sec:predict}
We next ask whether pre-attack \jocc{} differences forecast how readily forgotten knowledge recovers under relearning.

\paragraph{Model-level prediction.} We first evaluate whether pre-attack \jocc{} predicts recovery after relearning. For each evaluated unlearned model, we compute the Spearman correlation between its pre-attack accessibility and subsequent recovery after relearning.
Higher pre-attack \jocc{} is associated with greater recovery. 
As shown in Table~\ref{tab:predict}, across all attacked unlearned models, the Spearman correlation between accessibility and post-attack revival is $+0.35$. Higher-accessibility unlearned models also require substantially fewer fine-tuning steps to recover, with a negative Spearman correlation of $-0.70$. After controlling for behavioral forgetting and utility metrics, the association remains positive, with a correlation of $+0.33$.


The relationship between pre-attack accessibility and recovery is not explained solely by differences across unlearning algorithms. As shown in Table~\ref{tab:predict} and Fig.~\ref{fig:predict}, higher pre-attack \jocc{} consistently corresponds to greater recovery within method families, with positive within-method correlations across all eight methods. The pooled within-method correlation is $\rho=+0.45$ for identity-level accessibility and increases to $\rho=+0.71$ for the knowledge-level variant. These results indicate that \jocc{} captures checkpoint-level variation in relearning vulnerability rather than merely identifying weaker unlearning algorithms. The trend remains positive among behaviorally successful unlearned models.

\paragraph{Item-level prediction.}
Although \jocc{} predicts recovery vulnerability at the checkpoint
level, it does not provide an item-level deletion certificate. We
therefore evaluate whether pre-attack accessibility can distinguish facts that later recover from those that remain suppressed. As shown in Table~\ref{tab:item}, item-level prediction remains close to random:
\jocc{} provides no reliable discrimination between recovered and
non-recovered items, and adding it to behavioral and membership-based signals yields negligible improvement. These results indicate that residual accessibility captures a model's overall susceptibility to relearning, but does not reveal which specific facts are likely to return.

\paragraph{Interpreting the granularity gap.}
This observe admits a natural interpretation. Prior localization studies suggest that fine-tuning-based unlearning disables a shared retrieval pathway rather than erasing individual stored facts \citep{hong2024dissecting}. Under this view, relearning on a few entities reopens the common pathway, so which specific held-out items revive is governed largely by attack-time dynamics rather than by per-item residual traces present before the attack. Residual accessibility then behaves as a property of the checkpoint as a whole, quantifying how far the retrieval pathway has been displaced, which explains why it forecasts aggregate recovery while remaining uninformative about the fate of any single fact.

\begin{table}[t]
\centering
\small
\begin{tabular}{lrrrr}
\toprule
 & \jocc{} $\downarrow$ & UDS & MU & revival $\downarrow$ \\
\midrule
\multicolumn{5}{l}{\emph{Readout suppression }} \\
WD-Train $\lambda{=}0$  & 0.67 & 0.75 & 0.53 & 0.283 \\
WD-Train $\lambda{=}5$  & 0.57 & 0.78 & 0.48 & 0.347 \\
WD-Train $\lambda{=}10$ & 0.55 & 0.76 & 0.47 & 0.387 \\
\midrule
\multicolumn{5}{l}{\emph{Causal-criterion families}} \\
GradDiff (deep)   & 0.09 & 0.97 & 0.56 & 0.025 \\
RMU (deep)        & 0.05 & 0.99 & n/a  & 0.000 \\
GradDiff (shallow)& 0.62 & 0.34 & 0.44 & 0.800 \\
RMU (shallow)     & 0.59 & 0.17 & n/a  & 0.695 \\
IdkDPO  & 0.15 & 0.81 & 0.57 & 0.020 \\
\bottomrule
\end{tabular}
\caption{\textbf{Lower \jocc{} does not necessarily indicate genuine deletion.}
Increasing the WD-Train penalty lowers \jocc{} but increases post-attack revival without improving causal deletion depth. In contrast, GradDiff and RMU models with deeper deletion under UDS exhibit substantially less revival. MU denotes Model Utility; arrows indicate preferred directions.}
\label{tab:goodhart}
\end{table}

\subsection{Test 3: Should J-Access Be Used as an Unlearning Objective?}
\label{sec:goodhart}

The previous experiment establishes that \jocc{} is useful as an independent diagnostic: it reveals residual accessibility hidden by behavioral metrics and predicts model-level recovery under relearning attacks. This raises a natural but more difficult question. If residual accessibility is associated with recovery risk, can reducing \jocc{} itself produce more robust unlearning?
This question is important because internal metrics are often treated not only as evaluation tools but also as optimization objectives. We therefore test whether \jocc{} remains a reliable indicator when it becomes a training target. A valid deletion objective should reduce \jocc{} together with post-attack recovery.

\begin{figure}[t]
\centering
\includegraphics[width=\linewidth,]{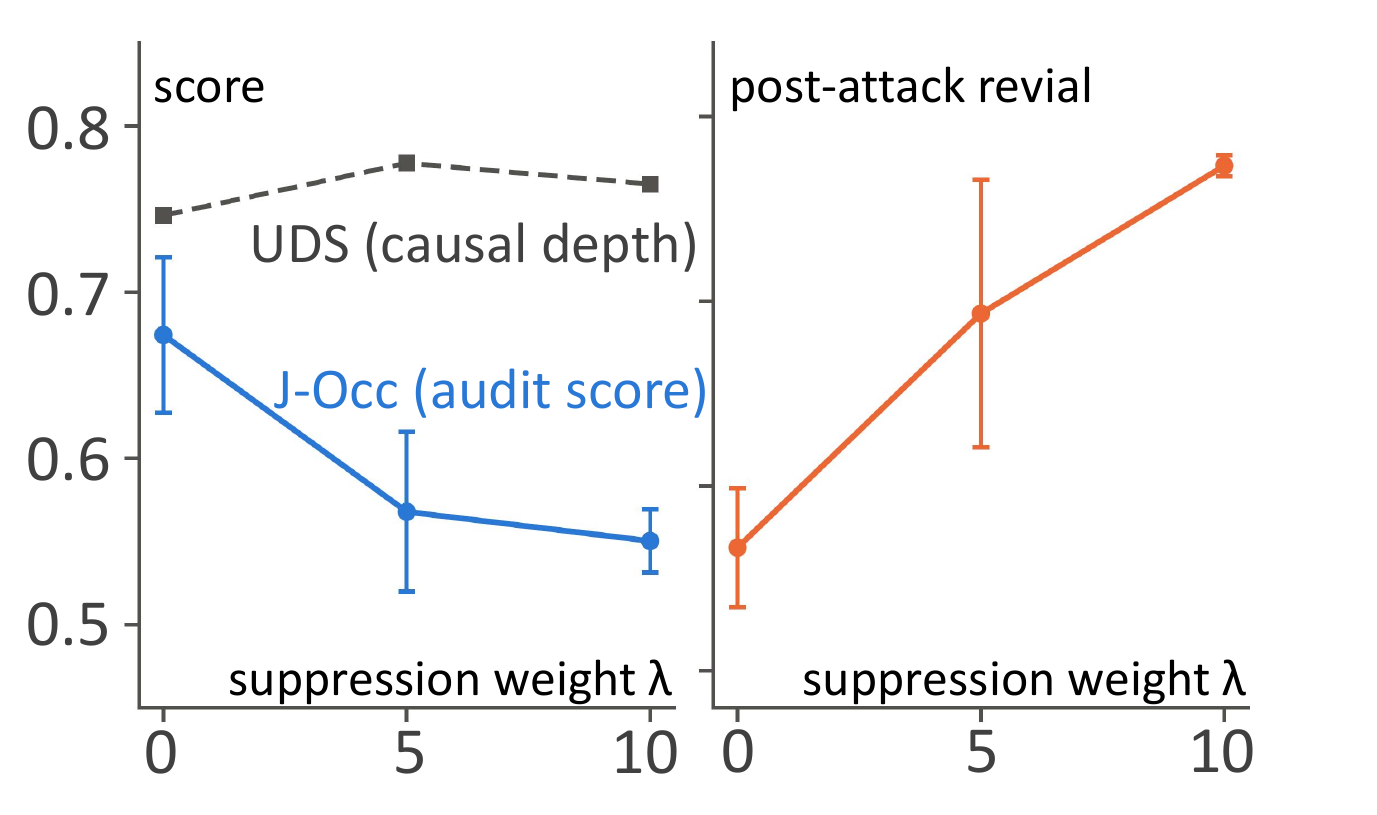}
\caption{\textbf{Directly minimizing \jocc{} suppresses the audit rather than deleting knowledge.}
As the penalty weight increases, \jocc{} decreases while UDS remains nearly unchanged (left), and post-attack revival increases (right).}
\label{fig:dose}
\end{figure}

Direct optimization produces the opposite of the expected pattern. As shown in Figure~\ref{fig:dose} and Table~\ref{tab:goodhart}, for WD-Train, increasing the suppression weight from $\lambda=0$ to $\lambda=10$ lowers \jocc{} from $0.67$ to $0.55$, but increases post-attack revival from $0.283$ to $0.387$. Thus, improvement on the optimized audit does not translate into more stable deletion, but instead coincides with greater vulnerability to subsequent knowledge recovery under attack.

To distinguish metric failure from a general failure of optimization, we compare accessibility suppression against unlearned models whose deletion is supported by an independent causal criterion. Models with deeper causal deletion consistently exhibit greater resistance to relearning. GradDiff and RMU configurations with high UDS achieve post-attack revival rates of only $0.025$ and $0.000$, respectively, whereas their shallow-deletion counterparts reach $0.800$ and $0.695$. Thus, UDS and recovery robustness move together across naturally occurring unlearning configurations, showing that the attack is capable of distinguishing genuine deletion from temporary suppression.

Accessibility suppression produces a different pattern: \jocc{} decreases while UDS remains nearly unchanged and post-attack recovery increases. Thus, optimizing \jocc{} suppresses evidence exposed to the audit without removing the knowledge that supports recovery. This asymmetry shows that a useful independent diagnostic is not necessarily a valid unlearning objective, which must be validated against causal deletion and recovery outcomes.

\section{Conclusion}
This work introduces \jocc{} to diagnose residual knowledge accessibility after machine unlearning. Experiments across 398 checkpoints and eight unlearning methods show that behavioral forgetting conceals internal access, and that pre-attack \jocc{} predicts recovery at the checkpoint level. We further show that predictive validity does not extend to individual facts, as item-level performance remains near chance. Although \jocc{} serves as an independent diagnostic, optimizing it suppresses the measured signal without producing genuine deletion and may increase recovery risk. Overall, reliable unlearning evaluation requires internal audits to complement behavioral metrics and be validated against independent causal and recovery evidence, rather than treated as deletion certificates or optimization targets.



\appendix

\bigskip

\bibliography{aaai2027}


\end{document}


\maketitle

\appendix


\paragraph{Limitation}

Our prospective validation supports \jocc{} only as a checkpoint-level
signal: it ranks unlearned models by how readily they recover, but it
does not identify which forgotten facts will return and licenses no
per-item deletion guarantee. The audited grid also shares a single
backbone family, and our optimization study instantiates one form of
audit suppression at three penalty weights, so the Goodhart effect we
report is an existence proof that lowering \jocc{} can be decoupled
from genuine deletion rather than evidence that every objective built
on an internal signal fails this way.

All experiments use TOFU, and this is not only a matter of compute.
Eq. 3 is defined by normalization against a retain-only
gold model $\theta_g$ that is identical to $\theta_0$ except that it
never observed the forget data. Real-world benchmarks such as WMDP
\citep{li2024wmdp} and RWKU \citep{cao2024rwku} target knowledge
acquired during pretraining, so no such counterpart exists and none can
be constructed without pretraining from scratch on a filtered corpus.
Without $\theta_g$ the zero point of the audit is undefined, and a
nonzero access rate cannot be separated from the background rate of a
model that never learned the target. The cost is external validity:
accessibility on TOFU may partly reflect memorization induced by the
fine-tuning that injected the forget data, and the reported magnitudes
should not be assumed to transfer. Extending the audit to real-world
knowledge first requires replacing $\theta_g$ with an independently
verified null condition, which we regard as a measurement-design
problem rather than a matter of running the same pipeline on more data.

\section{J-Access Construction and Implementation}
\label{app:jaccess}

\subsection{Jacobian-Lens Fitting}
\label{app:lens-fit}

We fit one Jacobian lens on the original model $\theta_0$ using the reference
implementation of \citet{gurnee2026verbalizable}. The corpus-averaged
Jacobians $J_\ell$ of Eq.~(1) in the main text are estimated for source layers
$0$--$14$ with the final layer as target, over 1{,}000 disjoint 128-token
chunks of WikiText-103, matching the corpus recipe of the original work.
Fitting takes roughly 72 minutes on a single GPU. The lens is then frozen and
shared across all audited checkpoints: each checkpoint contributes its own
residual states, final normalization, and unembedding in Eq.~(2), while the
transport matrices are held fixed. The audit is therefore identical in cost to
a forward pass with linear readouts, and no per-checkpoint fitting is required.
Robustness to this design choice---refitting the lens on unlearned models---is
evaluated in Appendix~\ref{app:validation}.

\subsection{Identity-Level Probes}
\label{app:probes-id}

Identity-level probes are drawn from the released
\texttt{forget10\_perturbed} split. For each forget author we keep questions
that satisfy two filters: no piece of the author's name appears in the question
(checked case-insensitively over all space- and hyphen-separated name parts
longer than two characters), and the reference answer states the name, so that
the name is exactly what should enter the workspace at answer onset. Three
sources pass these filters: the author's identity question, its released
paraphrase, and, where the identity question parses cleanly, a templated
reformulation of the form ``\emph{An author was born in $\langle$place$\rangle$
on $\langle$date$\rangle$. What is this author's full name?}''. This yields 55
forget probes with at least one probe per author. A matched set of 25
retain-author probes is constructed by the identical procedure from
\texttt{retain\_perturbed} and serves as the within-model comparison set.

\subsection{Identity Concept Tokens}
\label{app:concept-id}

The concept set $C_i$ of a probe collects the subword tokens of the author's
full name and of each name part, in both leading-space and bare tokenizations.
We keep tokens whose decoded string is at least three characters long, which
removes generic one- and two-letter BPE heads shared across many words, and
additionally keep the first token of each leading-space variant even when
short, because the model's first emitted token for a name can be such a head
(e.g., ``\texttt{ H}'' for \emph{Hsiao}). For example, \emph{Hsiao Yun-Hwa}
contributes the subword tokens of \emph{Hsiao}, \emph{Yun-Hwa}, \emph{Yun},
and \emph{Hwa}. The same construction, applied with each checkpoint's shared
tokenizer, is used for all models.

\subsection{Readout Positions and Workspace Band}
\label{app:positions}

Each probe is formatted with the OpenUnlearning TOFU chat template, and the
teacher-forced answer prefix ``\emph{The author's full name is}'' is appended.
The primary readout positions $\mathcal{P}$ are the tokens of this prefix,
where the name is about to be emitted; question and template positions are
recorded for diagnostics but do not enter the score. The search volume of the
audit is thus fixed and identical for every model, so differences in access
rate cannot arise from differences in the number of audited locations.

The workspace band $\mathcal{B}$ contains the four lens layers with the
highest full-model hit rate, computed on a calibration split of authors and
frozen before any unlearned checkpoint was audited: layers 11--14 at the
identity level and layers 10--13 at the knowledge level. The layerwise profile
in Appendix~\ref{app:validation} confirms that the readout is empty outside
the mid-to-late range for every model.

\subsection{Silent Set and Scoring}
\label{app:scoring}

A probe counts as accessible if the minimum rank of any concept token over
$\mathcal{B}\times\mathcal{P}$ is below $k$; $k=10$ is the preregistered
primary threshold and $k=50$ the secondary one. The silent set
$\mathcal{I}(\theta)$ is determined behaviorally: we decode a greedy
continuation for each probe and drop probes whose generated text contains any
concept token, since visible expression is the province of behavioral metrics,
not of the internal audit. Both anchors in Eq.~(6) are evaluated on the
audited checkpoint's own silent set. On the full probe set the empirical
anchors are $A_{10}(\theta_0)=0.582$ and $A_{10}(\theta_g)=0.291$; scores are
not clipped, so values outside $[0,1]$ indicate checkpoints beyond an anchor.

\subsection{Knowledge-Level Extension}
\label{app:concept-k}

The knowledge-level variant audits the original TOFU questions themselves: 398
forget and 398 retain probes (rows whose concept set is empty after filtering
are removed), plus 20 questions about fabricated twin authors
(Appendix~\ref{app:validation}). The concept set $K_i$ collects the content
words of the reference answer: alphabetic words of length at least four,
non-stopword, normalized for possessives, whose lowercase form does not occur
as a substring of the question (echo filter), and whose document frequency
across all 4{,}000 full-model TOFU answers is below 5\% (inferability filter,
removing generic filler words); if no word survives, the three lowest-frequency
content words are used. Each surviving word maps to the first subword token of
its leading-space form, subject to the same three-character rule; name tokens
are added only when no name piece occurs in the question. Readout uses nine
positions---the final prompt position and the first eight greedy generation
steps, re-reading the residual stream at each step---and the item score is
$-\log_{10}(\text{min rank}+1)$. Matched decoy sets pair every probe with the
$K$ set of the same-position question of a different author under a fixed
derangement of author indices, giving concept sets with the same construction
and distribution but no true association to the prompt.

\section{Validation Controls and Robustness Analyses}
\label{app:validation}

\paragraph{Echo controls.}
Three independent mechanisms rule out input copying. First, all probes are
name-absent by construction, and probes on which the model behaviorally emits
a concept token are excluded (Appendix~\ref{app:scoring}). Second, at the
knowledge level, concept words that occur in the question are removed from
$K_i$ before measurement. Third, the layerwise profile provides a direct
check: for every audited model the readout at layers 0--8 is identically zero;
the signal emerges from layer 9 and accumulates inside the workspace band,
where the separation between $\theta_0$ and $\theta_g$ concentrates. A readout
that copied the prompt would fire at early layers.

\paragraph{Unseen twin entities.}
We fabricate 20 twin authors whose names are verified by string search to
occur nowhere in TOFU, and pose identity questions of the same form as the
real probes. Measured access for these twins is $\approx 0.00$ at $k=10$ for
every model, including $\theta_0$. The audit's false-positive floor on
never-trained associations is therefore negligible.

\paragraph{Matched decoys.}
For each checkpoint we compare the ranks of the true concept set against the
matched decoy set of Appendix~\ref{app:concept-k} with a paired one-sided
Wilcoxon test on the same prompts. Only access in excess of the decoy baseline
is treated as evidence of a target-specific trace.

\paragraph{Full-grid inference.}
The per-checkpoint claim that raw access exceeds the gold anchor is tested
with a paired one-sided Wilcoxon test on $\log_{10}(\text{rank}+1)$ over
shared silent probes, Holm-corrected across all 398 checkpoints: 338 exceed
the gold anchor and 289 remain significant after correction. Raw access also
tracks behavioral forgetting dose-dependently, $\rho(\log_{10}\mathrm{FQ},
\text{access})=-0.49$ ($p=3.6\times10^{-25}$).

\paragraph{Lens refitting after unlearning.}
Because unlearning could shift representations in a way that invalidates a
lens fitted on $\theta_0$, we refit the lens from scratch, with the identical
corpus and recipe, on the highest-FQ checkpoint of each of the eight methods.
The paired effect direction is unchanged in 8 of 8 cases, and checkpoint
rankings under the refitted and shared lenses agree with $\rho=0.91$. The
shared-lens results reported in the paper are therefore not an artifact of
representation shift.

\paragraph{Convergent validity with UDS.}
On 36 checkpoints stratified by method and forgetting quality we compute the
Unlearning Depth Score~\citep{lee2026measuring}, a two-stage
activation-patching measure oriented so that 1 denotes deletion to gold depth.
Because \jocc{} measures residual access, the expected association is
negative: we observe $\rho=-0.72$ ($p=6.4\times10^{-7}$) at the identity level
and $\rho=-0.52$ ($p=0.0012$) at the knowledge level. Two mechanistically
distinct instruments---vocabulary-space decoding and causal patching---thus
order the same checkpoints consistently.

\paragraph{What behavioral metrics miss.}
Among 43 checkpoint pairs matched on Forget Quality, the 90th percentile of
the within-pair difference in \jocc{} is 0.51---the size of the entire
full-to-gold span. Matching models behaviorally leaves their internal
accessibility essentially unconstrained, which is the premise of the audit.

\section{Relearning Attacks and Prediction Analyses}
\label{app:relearning}

\subsection{Checkpoint Pool}
\label{app:pool}

The relearning study attacks 72 checkpoints: the measurement pool of
Appendix~\ref{app:validation} plus a supplement drawn by a fixed rule frozen
before any attack (per method, the epoch-10 candidates at grid corner and
midpoints, of which the minimum-, median-, and maximum-FQ candidates are
kept). The pool spans all eight methods and stratifies forgetting quality
within each method, and includes behaviorally unsuccessful checkpoints so that
correlations are not restricted to a narrow quality range.

\subsection{Attack Protocol}
\label{app:attack}

The primary attack is cross-entity partial relearning. The 20 forget authors
are split once (fixed seed) into 10 relearn authors and 10 held-out authors.
Each checkpoint is fine-tuned on all 200 question--answer pairs of the relearn
authors with the OpenUnlearning fine-tuning trainer (AdamW, effective batch
32, weight decay 0.01, bf16) for one epoch of 7 optimizer steps, and the 200
held-out questions---never shown during the attack---are evaluated after every
step. Recovery on held-out authors can therefore only arise from residual
structure, not from re-exposure. The learning rate $2\times10^{-5}$ was
selected from $\{1,2,5\}\times10^{-5}$ on four pilot checkpoints by a rule
fixed in the preregistration (maximize between-model variance of held-out
excess recovery, subject to the gold model recovering less than half of the
items), and then frozen. The gold model undergoes the identical attack, and
all outcomes are reported in excess of it.

Three checks support the protocol. A placebo fine-tune of the same budget on
retain-side data inflates the raw probability gain to $+0.167$, against
$+0.225$ for the real attack; most of the raw probability movement is thus
format drift, which motivates the ROUGE-based revival outcome below rather
than raw $\Delta$Prob. A second attack seed on a 24-checkpoint subset
reproduces outcomes with ICC $=0.9998$. Finally, 24 representative checkpoints
(three per method, stratified by within-method \jocc{}) were attacked for five
epochs with per-epoch archiving; the model-level correlation is stable across
budgets ($\rho$ between $+0.40$ and $+0.46$ at every epoch), so the reported
associations are not an artifact of the one-epoch horizon.

\subsection{Recovery Outcomes}
\label{app:outcomes}

An item counts as \emph{revived} if its post-attack ROUGE-L recall reaches
half of the original model's per-item score, having been below this threshold
before the attack; items already above threshold pre-attack are excluded from
item-level analyses. \emph{Excess revival} is a checkpoint's revival rate
minus that of the identically attacked gold model. \emph{Steps-to-recover} is
the first optimizer step at which mean held-out answer probability reaches
half of the original model's level; runs that never reach it are censored at
the horizon and enter the rank correlation at the maximum value.

\subsection{Model-Level Statistics}
\label{app:model-stats}

Model-level associations are one-sided Spearman correlations between
pre-attack \jocc{} and the outcomes above. The partial estimate is a partial
rank correlation controlling for Forget Quality and Model Utility. The
within-method estimate computes the correlation separately inside each of the
eight method families and pools the coefficients by inverse-variance weighting
after Fisher transformation; at the knowledge level the eight within-method
coefficients are $+0.27$ (AltPO), $+0.40$ (GradDiff), $+0.87$ (IdkDPO),
$+0.23$ (IdkNLL), $+0.77$ (NPO), $+0.93$ (RMU), $+0.88$ (SimNPO), and $+0.62$
(UNDIAL)---positive in every family. The preregistered strict primary
subset (Forget Quality $>0.01$) contains only 11 checkpoints and yields
$\rho=+0.18$, not significant; we report this as specified and rest the
conclusion on the full pool, the partial estimate, and the within-method
pooling, all fixed in the same preregistration.

\subsection{Item-Level Predictors}
\label{app:item-pred}

All predictors are measured before the attack. \emph{Pre-attack probability}
and \emph{pre-attack ROUGE} are the official per-item TOFU metrics. The six
membership-inference attacks are the OpenUnlearning implementations of loss,
zlib-entropy, Min-K\%, Min-K\%++, reference-model, and gradient-norm attacks.
The \emph{logit-lens} baseline computes the knowledge-level item score of
Appendix~\ref{app:concept-k} with the transport matrix replaced by the
identity, isolating the contribution of Jacobian transport. Performance is the
per-checkpoint AUROC for discriminating revived from non-revived eligible
items (checkpoints with at least 20 eligible items), summarized by the median
over 65 checkpoints. In the descriptive comparison of Table~3 (main text)
every predictor, including the MIA family from which the best of the six is
reported, is oriented in its empirically favorable direction; the
\emph{preregistered} row evaluates the same knowledge-level score with its
orientation fixed in advance (higher access $\Rightarrow$ revival) and no such
freedom, yielding a median AUROC of 0.504 (Wilcoxon versus 0.5, $p=0.31$).
The gap between the two rows measures exactly the optimism introduced by
post-hoc orientation.

\subsection{Incremental Value}
\label{app:stacked}

The stacked model is a logistic regression over standardized base predictors
(pre-attack probability, pre-attack ROUGE, a free-energy score, and the six
membership-inference scores), with and without the knowledge-level \jocc{}
feature. Evaluation uses five-fold cross-validation grouped by checkpoint, so
no checkpoint contributes to both training and test folds, and we report the
change in held-out AUROC: $+0.0002$ over a base of 0.682.

\section{Direct Optimization of J-Access}
\label{app:optimization}

\subsection{Training Objective}
\label{app:wdtrain-obj}

WD-Train augments a standard unlearning objective with an accessibility
penalty evaluated exactly where the audit reads:
\begin{equation}
\mathcal{L}
=
\gamma\,\mathcal{L}_{\mathrm{forget}}
+
\alpha\,\mathcal{L}_{\mathrm{retain}}
+
\lambda\,\mathcal{L}_{\mathrm{occ}},
\label{eq:wdtrain}
\end{equation}
where $\mathcal{L}_{\mathrm{forget}}$ is the NPO
loss~\citep{zhang2024negative}, $\mathcal{L}_{\mathrm{retain}}$ is the
retain-set cross-entropy, and the accessibility term is a margin hinge on the
lens log-probability of the ground-truth answer tokens:
\begin{equation}
\begin{split}
\mathcal{L}_{\mathrm{occ}}
=
\operatorname*{mean}_{\ell\in\mathcal{B},\, p\in\mathcal{A}}
\max\!\Big\{0,\;
&\log\operatorname{softmax}\!\big(
W_U \operatorname{Norm}(J_\ell h_{\ell,p})
\big)_{t_p} \\
&- m
\Big\},
\end{split}
\label{eq:occ}
\end{equation}
with $\mathcal{B}=\{10,\ldots,13\}$ the audited band, $\mathcal{A}$ the
answer-token positions of the forget batch, $t_p$ the true answer token at
position $p$, and margin $m=-6$. The top-$k$ indicator of the audit is not
differentiable; Eq.~\eqref{eq:occ} is its natural surrogate, driving the
lens-decoded probability of the tokens the audit searches for below a fixed
level rather than without bound. The transport matrices $J_\ell$ are the
frozen full-model lens and receive no gradient; gradients flow through the
residual states $h_{\ell,p}$ into the weights. The supervision targets are the
answer tokens the dataset already provides, so no audit probe or concept set
enters training.

\subsection{Training Setup}
\label{app:wdtrain-setup}

All runs use the OpenUnlearning trainer on TOFU \texttt{forget10} with the
retain split of $\theta_g$: NPO with $\beta=0.1$, $\gamma=\alpha=1$, learning
rate $5\times10^{-5}$, five epochs, effective batch size 32, AdamW, bf16. The
suppression weight takes $\lambda\in\{0,5,10\}$ with three random seeds per
setting; $\lambda=0$ is the matched NPO baseline, identical in every other
respect, so the effect of the penalty is isolated by construction. No setting
was selected post hoc: all $(\lambda,\text{seed})$ runs are reported.

\subsection{Evaluation and Separation}
\label{app:wdtrain-eval}

Each run is evaluated on four axes: Forget Quality and Model Utility
(official suite), knowledge-level \jocc{}, causal deletion depth
(UDS), and post-attack revival. The attack is the within-entity variant of the
frozen relearning protocol: for each held-out author, half of the questions
are relearned (learning rate $2\times10^{-5}$, six optimizer steps) and
revival is measured on the other half. The audit probes and concept sets are
never used in training (Appendix~\ref{app:wdtrain-obj}), and the attack
protocol and success criteria were frozen before the training grid was run.
Table~\ref{tab:wdtrain-seeds} reports seed-level results; the dose-dependent
reversal replicates in all three seeds, and a triple-budget attack preserves
it ($0.460$ at $\lambda=10$ versus $0.435$ at $\lambda=0$). The utility drop
from $\lambda=0$ to $\lambda=10$ is $0.06$ and cannot account for the revival
increase.

\begin{table}[t]
\centering
\small
\begin{tabular}{lccc}
\toprule
& $\lambda{=}0$ & $\lambda{=}5$ & $\lambda{=}10$ \\
\midrule
\jocc{}             & 0.674 & 0.568 & 0.550 \\
UDS                 & 0.75  & 0.78  & 0.76  \\
MU                  & 0.527 & 0.478 & 0.467 \\
revival             & $0.283_{\pm.016}$ & $0.347_{\pm.036}$ & $0.387_{\pm.003}$ \\
excess $\Delta$Prob & $+0.067$ & $+0.097$ & $+0.118$ \\
\bottomrule
\end{tabular}
\caption{...}
\label{tab:wdtrain-seeds}
\end{table}
\subsection{Comparison Models and Controls}
\label{app:wdtrain-controls}

The deep- and shallow-deletion comparisons in Table~4 (main text) are drawn
from the audited grid by their UDS values, selected to span the causal
criterion rather than matched to WD-Train: GradDiff at
$\mathrm{lr}=5\times10^{-5}$, $\alpha=10$ (UDS 0.97) versus
$\mathrm{lr}=1\times10^{-5}$, $\alpha=1$ (0.34), and RMU targeting layer 5
(0.99) versus layer 15 (0.17), with IdkDPO at $\mathrm{lr}=2\times10^{-5}$,
$\beta=0.05$ (0.81) as an intermediate refusal-trained point. All undergo the
identical within-entity attack. Across the four causal-family points, UDS and
post-attack robustness are perfectly rank-aligned, whereas WD-Train moves the
audit score and robustness in opposite directions while UDS stays flat.

Finally, to test whether the reversal is specific to the gradient-based
penalty, we also suppress the readout by an orthogonal mechanism: WD-Ablate
projects concept directions out of the weights (rank-32 parameter-space
projection) with no training signal at all. It drives identity-level \jocc{}
to 0.07 while revival is unchanged at 0.46. Two mechanistically unrelated ways
of optimizing the readout thus fail in the same way, supporting the
interpretation that the failure belongs to the act of optimizing the
observational signal, not to one particular training recipe.

\bigskip

\bibliography{aaai2027}